\documentclass[runningheads]{llncs}

\usepackage[T1]{fontenc}
\def\doi#1{\href{https://doi.org/\detokenize{#1}}{\url{https://doi.org/\detokenize{#1}}}}
\usepackage{caption}
\usepackage[final]{graphicx}
\usepackage{fancyhdr}
\usepackage{amssymb}
\usepackage{amsmath}
\usepackage{algorithm}
\usepackage{algpseudocode}
\usepackage{subcaption}
\usepackage{tabularx}  
\usepackage{booktabs}  
\usepackage{placeins}
\usepackage{cite}
\usepackage{svg}
\usepackage{hyperref}

\usepackage{xcolor}

\usepackage{listings}

\definecolor{codegreen}{rgb}{0,0.6,0}
\definecolor{codegray}{rgb}{0.5,0.5,0.5}
\definecolor{codepurple}{rgb}{0.58,0,0.82}
\definecolor{backcolour}{rgb}{0.95,0.95,0.92}

\lstdefinestyle{mystyle}{
    backgroundcolor=\color{backcolour},   
    commentstyle=\color{codegreen},
    keywordstyle=\color{magenta},
    numberstyle=\tiny\color{codegray},
    stringstyle=\color{codepurple},
    basicstyle=\ttfamily\footnotesize,
    breakatwhitespace=false,         
    breaklines=true,                 
    captionpos=b,                    
    keepspaces=true,                 
    numbers=left,                    
    numbersep=5pt,                  
    showspaces=false,                
    showstringspaces=false,
    showtabs=false,                  
    tabsize=2
}

\DeclareCaptionSubType*{algorithm}

\DeclareCaptionLabelFormat{alglabel}{Alg.~#2}
\usepackage{listings}
\begin{document}
\title{FrOoDo: Framework for Out-of-Distribution Detection}
%
%
\author{
Jonathan Stieber \and
Moritz Fuchs \and
Anirban Mukhopadhyay
}
\authorrunning{
}
%
\institute{
Technische Universität Darmstadt
}

\maketitle              
\begin{abstract}
FrOoDo is an easy-to-use and flexible framework for Out-of-Distribution detection tasks in digital pathology. It can be used with PyTorch classification and segmentation models, and its modular design allows for easy extension. The goal is to automate the task of OoD Evaluation such that research can focus on the main goal of either designing new models, new methods or evaluating a new dataset. The code can be found at \url{https://github.com/MECLabTUDA/FrOoDo}.

\end{abstract}

\section{Framework Design}

\begin{figure}[h]
\includegraphics[width=1\textwidth]{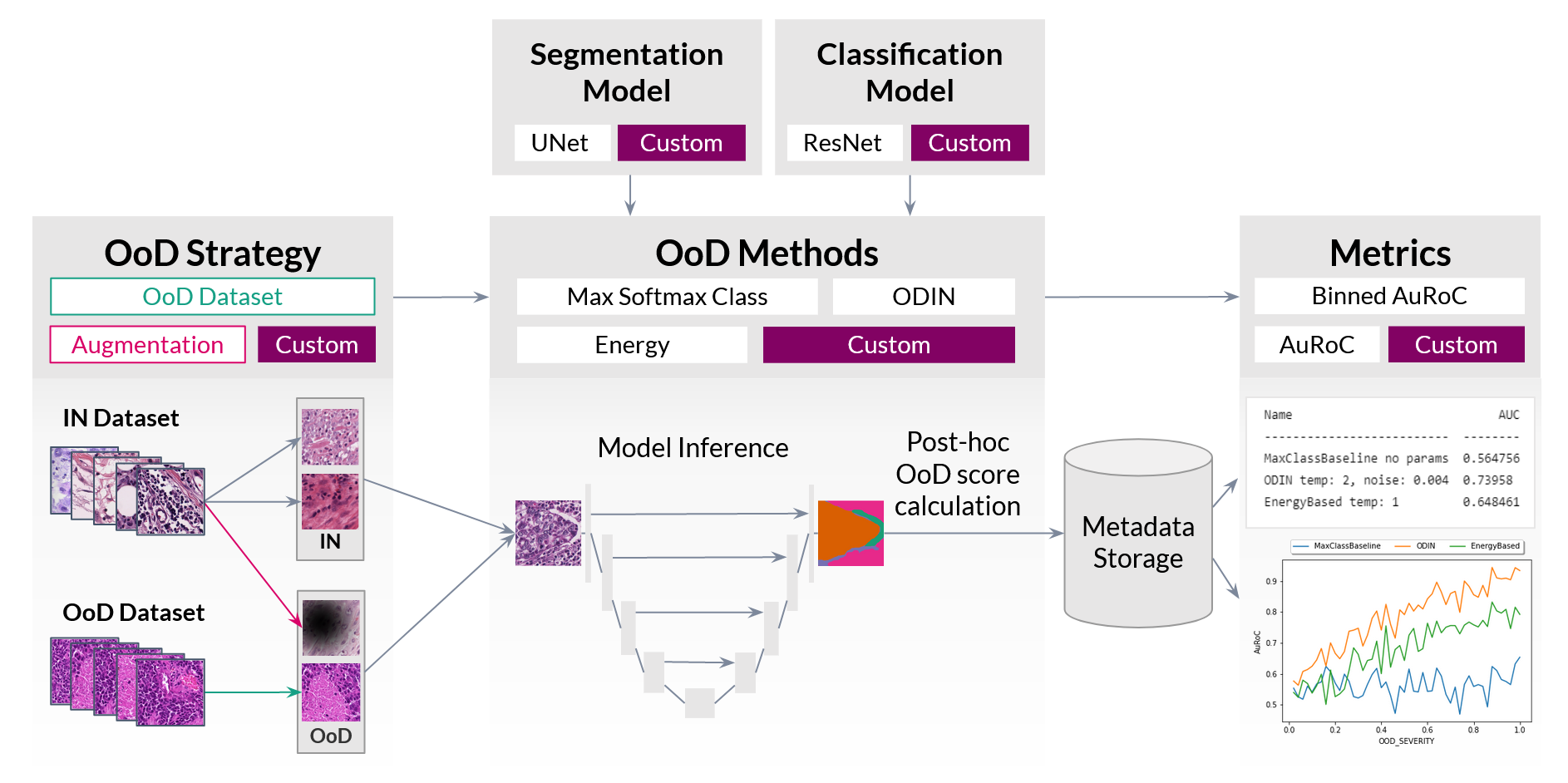}
\caption{Out-of-Distribution Evaluation Task Design}
\label{fig:overview}
\end{figure}

FrOoDo is a framework for \textbf{O}ut-\textbf{o}f-\textbf{D}istribution (OoD) detection tasks in digital pathology. It is created to help researchers in that area to quickly develop and evaluate new models, methods or datasets. It is designed to be easy-to-use, but also flexible to fit the requirements of researchers. The framework uses different building blocks that can be combined to form experiments and currently supports classification and segmentation models. That architecture, which can be seen in \ref{fig:overview}, allows researchers to modify or create new blocks regarding their research interest. Using a fixed architecture and only changing specific blocks ensures that results are reproducible and comparable.

\subsection{Out-of-Distribution Strategies}
The heart of the framework are the different OoD strategies, which determines what data is considered OoD data or in which way the OoD data is generated. Currently, the frameworks supports the augmentation strategy and the OoD dataset strategy.
\subsubsection{Augmentations}
One way of generating OoD data is by augmenting In-Distribution (IN) data in a task related way. Since FrOoDo is designed for digital pathology, it already contains pathology specific augmentations \cite{schomig2021quality}.

\begin{figure}[h]
\includegraphics[width=1\textwidth]{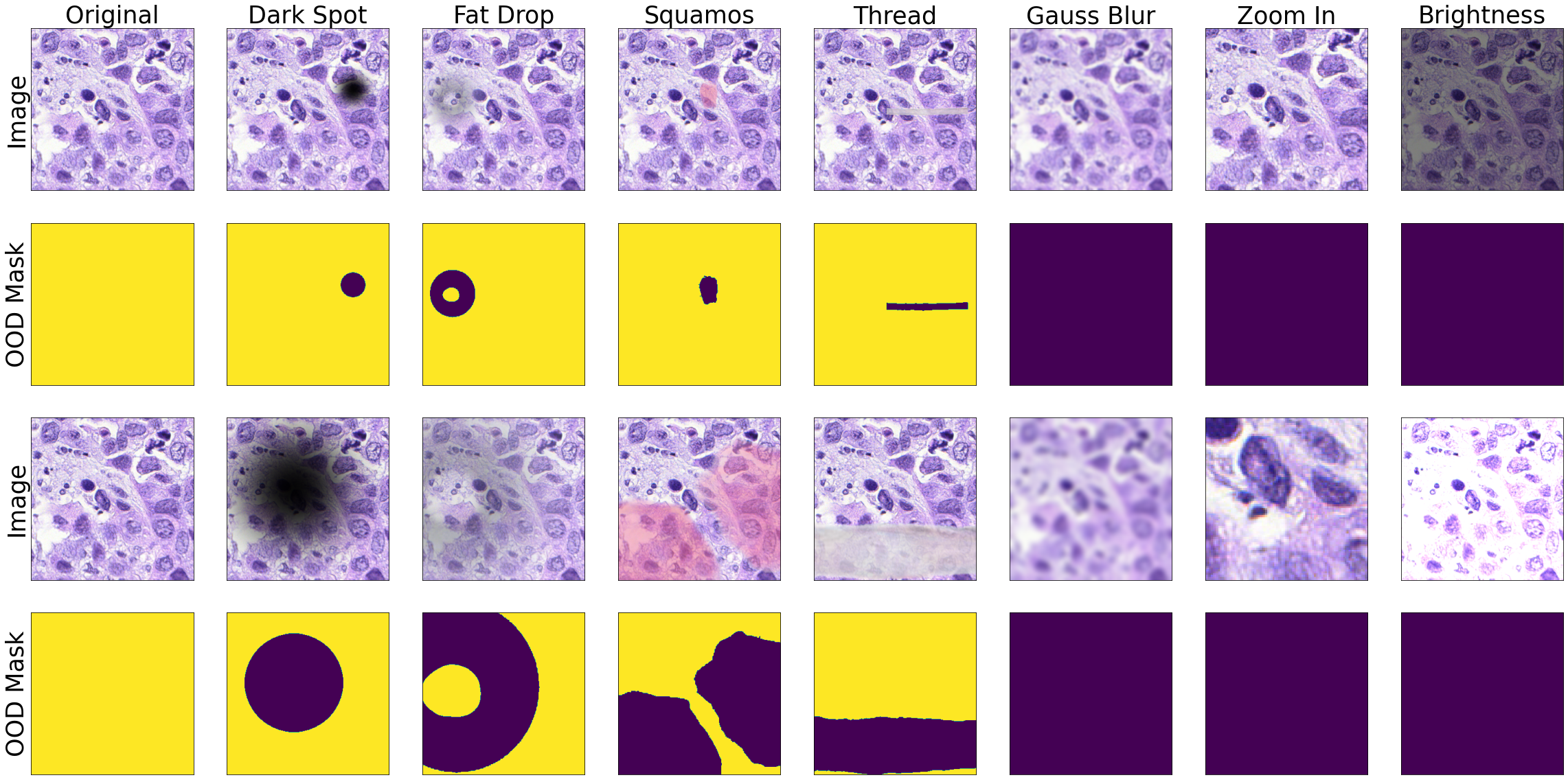}
\caption{Currently supported augmentations applied to an image of the BCSS dataset \cite{amgad2019structured} and the resulting OoD images and masks}
\label{fig:aug}
\end{figure}
As it can be seen in Figure \ref{fig:aug}, the augmentations give details about what part of the image can be seen as OoD. Therefore, the OoD evaluation can also be performed on subsets of the data w.r.t. the percentage of OoD pixels.
\subsubsection{Out-of-Distribution Dataset}
Beside automatically corrupting images of an IN dataset to receive OoD images, one can also take a whole dataset and mark it as OoD data. This is especially helpful if you want to compare different IN datasets with OoD datasets from e.g.  institutes with staining or other image acquisition differences.

\subsection{OoD Methods}
Currently, FrOoDo only supports post-hoc OoD detection methods. These methods do not require information about the network architectures and only use the output of the network for the OoD score.
FrOoDo already comes with well known methods implemented.

\subsubsection{Maximum Class Softmax Score~\cite{hendrycks2016baseline}} Taking the maximum value of the softmax distribution as the OoD score is the simplest and most intuitive method, since one would expect that the network gives lower probability for OoD images. However, most neural networks fail silently as they still produce high certain prediction for OoD data. Therefore, this method should only serve as a lower bound baseline.

\subsubsection{Energy~\cite{liu2020energy}}
Instead of only considering the maximum softmax value, Energy scoring uses the whole softmax distribution for the OoD score calculation.
This is done by calculating the logsumexp over the all softmax values.
$$ E(x;T) = - T * log \sum_{i}^K e^{-f_i(x)/T}$$
In general, this yields a better separability between In and Out-of-Distribution data.

\subsubsection{ODIN~\cite{liang2018enhancing}} Liang et al. apply image perturbation to the input by evaluating the softmax score against the maximum logit $\Tilde{y}$:
$$ \Tilde{x}=x - \epsilon sign(-\nabla_x log( S_{\Tilde{y}}(x;T))),$$
where the parameter $\epsilon$ is the perturbation magnitude. A perturbation can significantly decrease prediction accuracy in adversarial attacks. Applying a perturbation in this context corresponds to a severe decrease of the originally predicted class softmax score. The goal of applying image perturbation in this setting is to increase the softmax score for the maximum logit $\Tilde{y}$. The method assumes that the maximum logit is correctly predicted and uses the logit as a pseudo-label for the cross-entropy loss. Liang et al. show that this method has a greater effect on in-distribution images, leading to a better separation between in and out-of-distribution.

\section{Usage}

Using the framework is simple as you only have to choose the building blocks for the specific experiment which is the PyTorch network, an augmentation strategy, OoD detection methods, metrics and task type.

\begin{lstlisting}[language=Python]
from froodo import *
# init network
net = SegmentationModel().load()
# create BCSS dataset adapter (see BCSS docu)
adapter = GeneralDatasetAdapter(BCSS_Adapted_Cropped_Resized_Datasets().test)
# choose metrics
metrics = [OODAuRoC()]
# choose post-hoc OoD methods
methods = [MaxClassBaseline(), ODIN(), EnergyBased()]
# create experiment component
experiment = AugmentationOODEvaluationComponent(
    adapter,
    SampledAugmentation(DarkSpotsAugmentation()),
    model=net,
    metrics=metrics,
    methods=methods,
    seed=4321,
    task_type=TaskType.SEGMENTATION
)
# run experiment
experiment()
\end{lstlisting}

\subsection{Flexibility}

All the building blocks have clearly defined interfaces. If one wants to extend the framework, it is only necessary to fulfil the interface requirements.

\subsubsection{Dataset adaptation}
To make evaluation of the own datasets as easy as possible, FrOoDo uses dataset adapters to transform the data of any dataset into the data types it uses internally.

\begin{lstlisting}[language=Python]
# load your dataset
dataset = load_my_dataset()
# adapt dataset to FrOoDo 
adapted_dataset = GeneralDatasetAdapter(dataset, remapping=your_remapping)
\end{lstlisting}
Once a dataset is adapted, it can be used for any built-in operations and evaluation tasks. A detailed explanation about dataset adaptation can be found at \url{https://github.com/MECLabTUDA/FrOoDo/blob/main/docs/DATASET_ADAPTER.md}

\subsubsection{OoD method}
To create a new OoD method, the only mandatory requirement is to implement the calculate\_ood\_score method. For each image in a batch, the method should return pixel-wise values for segmentation and a single value for classification.
\begin{lstlisting}[language=Python]
# Method needs to inherit from OODMethod
class NewMethod(OODMethod):
    # only mandatory method for a new OoD method
    def calculate_ood_score(self, imgs, net, batch=None):
        # return OoD score
\end{lstlisting}

\subsubsection{Augmentation}
To implement new augmentations, it is again only necessary to implement the given Interfaces.
\begin{lstlisting}[language=Python]
class NewAugmentation(OODAugmentation):
    # only mandatory method to implement a new OoD augmentation
    def _augment(self, sample: Sample) -> Sample:
        # augment image and change OoD mask accordingly
\end{lstlisting}
Besides the shown augmentations, there are also special augmentations and augmentation interfaces that e.g. control the sampling of augmentation parameters. A more detailed overview on the available augmentation can be found at \url{https://github.com/MECLabTUDA/FrOoDo/blob/main/docs/AUGMENTATION.md}.

\subsection{Reproducibility and Comparability}

To ensure comparability between the OoD methods, during an experiment every method is presented exactly the same images. When loading in a batch of images, they are subsequently fed into the methods and the OoD score is calculated. Therefore, the framework ensures, that differences between method performances are not due to different image samples. 

To reduce the impact of randomness in the experiments, every experiment can be given a random seed. This seed ensures, that in all experiments with this seed, exactly the same images are chosen and generated by augmentations.

\bibliographystyle{splncs04}
\bibliography{bibliography}
\end{document}